\documentclass[table]{article}

\usepackage[table]{xcolor}
\usepackage{tikz}
\usetikzlibrary{shadows}
\usepackage{subcaption}

\usepackage[final]{neurips_2025}

\usepackage[utf8]{inputenc} 
\usepackage[T1]{fontenc}    
\usepackage{hyperref}       
\usepackage{url}    
\usepackage{booktabs}       
\usepackage{amsfonts}  
\usepackage{amsmath}
\usepackage{nicefrac}       
\usepackage{microtype}      

\def \TM{M_{\text{target}}}

\def \taskname{PropInfer}

\usepackage{algorithm}
\usepackage{multirow}
\usepackage{enumitem}
\usepackage{algpseudocode}

\title{Can We Infer Confidential Properties of Training Data from LLMs?}
\author{
   Pengrun Huang, Chhavi Yadav, Kamalika Chaudhuri$^{\dagger}$, Ruihan Wu$^{\dagger}$
}

\begin{document}
\footnotetext[1]{$^\dagger$ Equal advising.}
\maketitle
\vspace*{-0.7cm}
\begin{center}
    { University of California, San Diego} \\
\texttt{\{peh006, cyadav, kamalika, ruw076\}@ucsd.edu} %
\vspace*{0.5cm} 
\end{center}

\begin{abstract}
Large language models (LLMs) are increasingly fine-tuned on domain-specific datasets to support applications in fields such as healthcare, finance, and law. These fine-tuning datasets often have sensitive and confidential dataset-level properties — such as patient demographics or disease prevalence—that are not intended to be revealed. While prior work has studied property inference attacks on discriminative models (e.g., image classification models) and generative models (e.g., GANs for image data), it remains unclear if such attacks transfer to LLMs. In this work, we introduce \taskname, a benchmark task for evaluating property inference in LLMs under two fine-tuning paradigms: question-answering and chat-completion. Built on the ChatDoctor dataset, our benchmark includes a range of property types and task configurations. We further propose two tailored attacks: a prompt-based generation attack and a shadow-model attack leveraging word frequency signals. Empirical evaluations across multiple pretrained LLMs show the success of our attacks, revealing a previously unrecognized vulnerability in LLMs. We release our code at \href{https://github.com/PengrunH/Property_inference_attack_LLM}{github.com/PengrunH/Property\_inference\_attack\_LLM}.


%
%
\end{abstract}

\section{Introduction}
Large language models (LLMs) are increasingly deployed in real-world applications across domains such as healthcare \cite{he2025surveylargelanguagemodels}, finance \cite{li2024largelanguagemodelsfinance}, and law \cite{lai2023largelanguagemodelslaw}.  To adapt to domain-specific tasks, such as customer service or tele-medicine, these models are typically fine-tuned on proprietary datasets that are relevant to the tasks at hand before deployment. These domain-specific fine-tuning datasets however often contain \emph{dataset-level confidential information}. For example, a customer-service dataset sourced from a business may contain information about their typical customer-profile; a doctor-patient chat dataset sourced from a hospital may contain patient demographics or the fraction of patients with a sensitive disease such as HIV. Many businesses and medical practices would consider this kind of information non-public for business or other reasons. Thus, unintentional leakage of this information through a deployed model could lead to a breach of confidentiality.
Unlike individual-level privacy breaches that is typically addressed by rigorous definitions such as differential privacy~\citep{dwork2006calibrating,dwork2014algorithmic}, the risk here is the leakage of dataset-level properties.

Prior work has investigated this form of leakage, commonly referred to as \emph{property inference}~\citep{ateniese2013hackingsmartmachinessmarter}.
Most of the literature here has focused on two settings. 
The first involves discriminative models trained on tabular or image data~\citep{ateniese2013hackingsmartmachinessmarter, ganju2018property, chase2021propertyinferencepoisoning, suri2024dissectingdistributioninference, zhang2021leakagedatasetpropertiesmultiparty, 10136150}, where the goal is to infer attributes such as the gender distribution in a hospital dataset.
The second focuses on generative models~\citep{zhou2021propertyinferenceattacksgans, wang2024property}, such as GANs for face synthesis, where attackers may attempt to recover aggregate properties such as the racial composition of the training data.
In both cases, property inference has been shown to be feasible, and specialized attacks have been proposed to exploit these vulnerabilities. 

However, property inference in large language models (LLMs) introduces two distinct challenges. 
First, unlike inferring a single attribute from models trained on tabular data, the sensitive properties are more complex and may be indirectly embedded within the text.
For example, gender might be implied through broader linguistic cues, such as the mention of a “my gynecologist”.
LLMs may memorize such properties implicitly, making them more challenging to infer reliably in property inference studies.
The second challenge is that, unlike the models typically studied in prior work, LLMs do not fit cleanly into purely discriminative or generative categories; this raises questions about what kind of property inference attacks apply and succeed for these problems. 



In this work, we investigate both questions by introducing a new benchmark task -- \taskname \footnote{https://huggingface.co/datasets/Pengrun/PropInfer\_dataset} -- for property inference in LLMs. Our task is based on the Chat-Doctor dataset~\citep{li2023chatdoctormedicalchatmodel} -- a domain-specific medical dataset containing a collection of question-answer pairs between patients and doctors. There are two standard ways to fine-tune an LLM with this dataset that correspond to two use-cases: question-answering and chat-completion. According to the use case, our benchmark task has two modes where the models are fine-tuned differently -- Q\&A Mode and Chat-Completion Mode. To comprehensively study property inference across the two modes of models, we select a range of properties that are explicitly or implicitly reflected in both questions and answers.

We propose two property inference attacks tailored to LLMs.
The first is a black-box generation-based attack, inspired by prior work~\citep{zhou2021propertyinferenceattacksgans}; the intuition is that the distribution of the generated samples reflect the distribution of the fine-tuning data.
Given designed prompts that reflects characteristics of the target dataset, the adversary generates multiple samples from the target LLM and labels each based on the presence of the target property. The property ratio is then estimated by aggregating the labels.
The second is a shadow-model attack with word-frequency.
With access to an auxiliary dataset, the adversary first trains a set of shadow models with varying property ratios and extracts word frequency from the shadow models based on some selected keyword list. Then the adversary trains a meta-attack model that maps these frequencies to the corresponding property ratios.
This enables the inference on the target model by computing its output word frequencies.



We empirically evaluate our two attacks alongside baseline methods using our \taskname -benchmark.
Our results show that the shadow-model attack with word frequency is particularly effective when the target model is fine-tuned in the Q\&A Mode \textbf{and} the target property is more explicitly revealed in the question content than the answer.
In contrast, when the model is fine-tuned in Chat-Completion Mode \textbf{or} when the target attribute are embedded in both the question and the answer, the black-box generation-based attack proves to be simple yet highly effective.

Our experimental results reveal a previously underexplored vulnerability in large language models: property inference, which enables adversaries to extract dataset-level attributes from fine-tuned models. 
This finding exposes a tangible threat to data confidentiality in real-world deployments. 
It also underscores the need for robust defense mechanisms to mitigate such attacks -- an area where our benchmark provides a standardized and extensible framework for future research and evaluation.

\section{Related Work}


\textbf{Property inference.} Property Inference Attack (PIA) was first described by \cite{ateniese2013hackingsmartmachinessmarter}, as follows: given two candidate training data distributions $\mathcal{D}_1, \mathcal{D}_2$ and a target model, the adversary tries to guess which training distribution (out of $\mathcal{D}_1, \mathcal{D}_2$) is the target model trained on. Typically, the two candidate distributions only differ in the marginal distribution of a binary variable, such as gender ratio. A major portion of past work on property inference focuses on discriminative models \cite{ateniese2013hackingsmartmachinessmarter, ganju2018property, chase2021propertyinferencepoisoning, suri2024dissectingdistributioninference, zhang2021leakagedatasetpropertiesmultiparty, 10136150}; here the attacks mainly rely on training meta-classifiers on some representations to predict target ratio. For example, in the white-box setting, \cite{ateniese2013hackingsmartmachinessmarter, ganju2018property} use model weights as the input of the meta-classifier to predict the correct distribution. In the grey-box setting, where the adversary have access to the training process and some auxiliary data, \cite{suri2022formalizingestimatingdistributioninference, suri2024dissectingdistributioninference} use model outputs such as loss or probability vector as inputs to the meta-classifier. 

Moving on to generative models, \cite{zhou2021propertyinferenceattacksgans} study property inference attack for GANs.
The target GANs are trained on a human-face image dataset, whereas the adversary's task is to predict the ratio of the target property among the dataset, such as gender or race. Their attack follows the intuition that the generated samples from GANs can reflect the training distribution. Later on, \cite{wang2024property} studies property existence attacks. For example, if any images of a specific brand of cars are used in the training set. 

Contrary to previous works which either focus on discriminative models or pure generative models, we consider property inference attack for large language models. Since the model architecture, training paradigm and data type for LLMs are very distinct from previous works, it is unclear whether previous attacks still apply in the LLM setting.


\textbf{Other related works on data privacy and confidentiality in LLMs.} 
\cite{carlini2021extractingtrainingdatalarge} study training data extraction from LLMs, aiming to recover individual training samples. While one might try to infer dataset properties from extracted data, this often fails because the extracted samples are typically biased and not representative of the overall distribution. \cite{maini2024llmdatasetinferencedid} investigate dataset inference attacks, which aim to identify the dataset used for fine-tuning from a set of candidates. In contrast, our goal is to infer specific aggregate properties, not the dataset itself. \cite{sun2025idiosyncrasieslargelanguagemodels} study idiosyncrasies in public LLMs, determining which public LLM is behind a black-box interface. Although they also use word frequency signals, their objective differs fundamentally from ours.

\section{Preliminaries}
\subsection{Large Language Models Fine-Tuning} \label{LLM Preliminary}
A large language model (LLM) predicts the likelihood of a sequence of tokens.
Given input tokens $t_0, ... , t_{i-1}$, the language model parameterized by parameters $\theta$, $f_\theta$, outputs the distribution of the possible next token $f_\theta(t_i|t_0, ..., t_{i-1})$. 
Pre-training LLMs on large-scale corpora enables them to develop general language understanding and encode broad world knowledge.
In pre-training, the LLM is trained to maximize the likelihood of unlabeled text sequences.
Each training sample is a document comprising a sequence of tokens, and the objective is to minimize the negative log-likelihood:
$
    \mathcal{L}(\theta) = -\sum_{i=1}^k \log f_{\theta}(t_i|t_0,...t_{i-1})
$
 where $k$ is the total number of tokens in the document.

After pre-training, LLMs are often fine-tuned on domain-specific datasets to improve performance on downstream tasks.
The data in such datasets typically consists of an instruction  ($I$), which generally describes the task in natural language, and a pair of an input ($x$) and a ground-truth output ($y$). 
Two popular fine-tuning approaches are:
\begin{enumerate}[leftmargin=*,nosep]
\item \textbf{Supervised Fine-Tuning} (SFT;~\citep{radford2018improving, ouyang2022training}). SFT minimizes the negative log-likelihood of the output tokens conditioned on the instruction and input.
This approach focuses on learning the mapping from $(I, x)$ to $y$ and is commonly used in question-answering tasks.
The objective is: $\mathcal{L}_{\rm SFT}(\theta) = -\sum_{i=1}^{l}\log f_\theta(y_i|I,x, y_0,...,y_{i-1}).$

\item \textbf{Causal Language-Modeling Fine-Tuning} (CLM-FT; ~\citep{radford2018improving}). Different from SFT, CLM-FT follows the pre-training paradigm and minimizes the loss over all tokens in the concatenated sequence $t$ of instruction, input, and output $(I, x, y)$.
This method treats the full sequence autoregressively, making it suitable for tasks involving auto-completion for both user and chatbot.
The objective is $\mathcal{L}_{\rm CLM-FT}(\theta) = -\sum_{i=1}^k \log f_{\theta}(t_i|t_0,...t_{i-1}).$


\end{enumerate}

\subsection{Property Inference Attack}\label{subsec:attacksetup}
In this paper, we focus on LLMs that have been fine-tuned on domain-specific datasets, as these datasets often encompass scenarios involving confidential or sensitive information.
Given an LLM fine-tuned on such a dataset, property inference attacks aim to extract the dataset-level properties of the fine-tuning dataset from the finetuned LLM, which the data owner does not intend to disclose
\footnote{When the property is correlated with what the model learns, it seems pessimistic to avoid such leakage. However, the properties of concern in practice are often orthogonal to the task itself. 
See details in Section~\ref{sec:discuss}.}.

Let $\mathcal{S} = {(x_i, y_i)}_{i=1}^n$ denote the fine-tuning dataset of size $n$, consisting of i.i.d. samples drawn from an underlying distribution $\mathcal{D}$ over the domain $X \times Y$.
We denote the fine-tuned model as $f = \mathcal{A}(\mathcal{S}; I)$, where $\mathcal{A}$ is the fine-tuning algorithm applied to $\mathcal{S}$ using a fixed instruction template $I$.
Let $P: X \times Y \to \{0, 1\}$ be a binary function indicating whether a particular data point satisfies a certain property.
For example, $P(x, y) = 1$ may indicate that a patient in a doctor-patient dialogue $(x, y)$ is female.
The adversary's goal is to estimate the ratio of the target property $P$ among the dataset $S$.
$\text{The adversary's goal is: } r(P, S):=\frac{1}{n} \sum_{i=1}^n P(x_i, y_i).$

\subsection{Threat Models} \label{sub:threat models}
We consider two standard threat models in this work: black-box setting and grey-box setting.

\textbf{Black-box setting.} 
Following prior work~\citep{zhou2021propertyinferenceattacksgans}, we consider the black-box setting in which the adversary has only API-level access to the target model $f$.
In the LLM context, this means the adversary can create arbitrary prompts and receive sampled outputs from the model, but has no access to its parameters, architecture, or the auxiliary data.
This represents the most restrictive and least informed setting for the adversary, where only input-output interactions are observable.

\textbf{Grey-box setting.} Another standard threat model in the literature is this grey-box access~\citep{zhou2021propertyinferenceattacksgans, suri2022formalizingestimatingdistributioninference, suri2024dissectingdistributioninference}. 
In addition to black-box access to the target model $f$, we assume the adversary (1) has knowledge of the fine-tuning procedure $\mathcal{A}$, including details of the pre-trained model, fine-tuning method and the instruction template $I$, (2) has the knowledge of target dataset size $n$, and (3) has an auxiliary dataset $\mathcal{S}_{\text{aux}} = {(\hat{x}_i, \hat{y}_i)}_{i=1}^{n'}$ drawn i.i.d. from the underlying distribution $\hat{\mathcal{D}}$.
The inference problem becomes trivial when $\hat{\mathcal{D}}$ is the same as $\mathcal{D}$ where the fine-tuning dataset $\mathcal{S}$ is sampled from.
To make the setting nontrivial and realistic, we assume that $\hat{\mathcal{D}}$ and $\mathcal{D}$ differ only in the marginal distribution of the target property, while sharing the same conditional distribution given the property.
\section{\taskname: Benchmarking Property Inference Across Fine-Tuning and Property Types}\label{sec:benchmark}

We build our benchmarks upon a popular patient-doctor dialogues dataset ChatDoctor~\cite{li2023chatdoctormedicalchatmodel}; Figure \ref{fig:ChatDoctor dataset} shows examples of the dataset.
In this setting, an adversary may attempt to infer sensitive demographic attributes or the frequency of specific medical diagnoses — both representing realistic threats in which leakage of aggregate properties could have serious consequences.
To systematically study property inference attacks in LLMs, we extend the original ChatDoctor dataset by introducing two modes of the fine-tuned models, and the target properties, into our benchmark.

\begin{figure*}[t]
\centering
\begin{tikzpicture}

\def\cardwidth{5.5cm}
\def\cardheight{2.5cm}

\tikzstyle{cardbg} = [draw=black!30, fill=gray!3, rounded corners=2pt, text width=\cardwidth, minimum height=\cardheight, drop shadow]
\foreach \i in {2,1} {
    \node[cardbg] (f\i) at (0+\i*0.15,-\i*0.15) {};
}
\node[cardbg, drop shadow={shadow xshift=0.3ex, shadow yshift=-0.3ex}] (femaleCard) at (0,0) {
  \textbf{Patient:} My 3 year old \textcolor{red}{daughter} complains headache and throat pain since yesterday.What should I do?\\[0.3em]
  \textbf{Doctor:} The most likely reason for these symptoms are self limiting viral infections. I recommend...
};

\foreach \i in {2,1} {
    \node[cardbg] (d\i) at (7+\i*0.15,-\i*0.15) {};
}
\node[cardbg, drop shadow={shadow xshift=0.3ex, shadow yshift=-0.3ex}] (digestiveCard) at (7,0) {
  \textbf{Patient:} I often have \textcolor{red}{stomach cramps} and \textcolor{red}{diarrhea} after eating. It’s been going on for months now.\\[0.3em] 
  \textbf{Doctor:} This could be related to \textcolor{red}{irritable bowel syndrome}.
};

\node[draw=blue, fill=blue!10, font=\scriptsize, text centered, minimum width=1.2cm, anchor=center, rounded corners=2pt] (fl1) at (-0.5 + 0*1.4,-2) {Female=1};
\draw[->, blue!60!black] (fl1.north) -- ([xshift=-3.5mm,yshift=2.5mm] f1.south);

\node[draw=blue, fill=blue!10, font=\scriptsize, text centered, minimum width=1.2cm, anchor=center, rounded corners=2pt] (fl2) at (-0.5 + 1*1.4,-2) {Female=0};
\draw[->, blue!60!black] (fl2.north) -- ([xshift=-3.5mm,yshift=1mm] f2.south);

\node[draw=blue, fill=blue!10, font=\scriptsize, text centered, minimum width=1.6cm, anchor=center, rounded corners=2pt] (dl1) at (6 + 0*1.7, -2) {Digestion=1};
\draw[->, blue!60!black] (dl1.north) -- ([xshift=-3.5mm, yshift=2.5mm] d1.south);

\node[draw=blue, fill=blue!10, font=\scriptsize, text centered, minimum width=1.6cm, anchor=center, rounded corners=2pt] (dl2) at (6 + 1*1.7, -2) {Digestion=0};
\draw[->, blue!60!black] (dl2.north) -- ([xshift=-3.5mm, yshift=1mm] d2.south);

\node at (0,-2.5) {Female ratio = 70\%};
\node at (7.5,-2.5) {Digestion ratio = 12.68\%};

\end{tikzpicture}
\caption{This figure demonstrates examples of the ChatDoctor dataset and the property labels. (Left) An example of dialogue explicitly indicate the patient is a female, since it mentioned "daughter"; (right) an example of dialogue indicating the patient is consulting about digestive disorder.}
\vspace{-2ex}
\label{fig:ChatDoctor dataset}
\end{figure*}

\paragraph{Two modes of the fine-tuned models: Q\&A Mode and Chat-Completion Mode.} 
The ChatDoctor dataset supports two common use cases: (1) Doctor-like Q\&A chatbot for automatic diagnosis, and (2) Chat-completion to assist both patients and doctors.
Let $x$ denote the patient's symptom description and $y$ the doctor's diagnosis.
In the Q\&A chatbot mode, models are fine-tuned using \textit{Supervised Fine-Tuning} (SFT), learning to generate $y$ conditioned on $I$ and $x$.
In the chat-completion mode, models are trained using \textit{Causal Language-Modeling fine-tuning} (CLM-FT), which minimizes loss over the entire sequence of tokens in $I$, $x$, and $y$. This allows the model to predict tokens at any point in the dialogue. For the formal training objectives, kindly refer to Section \ref{LLM Preliminary}.
Accordingly, our benchmark includes both the Q\&A Mode and the Chat-Completion Mode, reflecting two widely used fine-tuning paradigms: SFT and CLM-FT.

These two modes naturally introduce different memorization patterns: CLM-FT encourages the model to learn the joint distribution $\mathbb{P}(I, x, y)$, potentially memorizing both patient and doctor texts equally; differently, SFT focuses on the conditional distribution $\mathbb{P}(y | I, x)$, emphasizing the doctor's response $y$ more heavily than the patient's input $x$.
Consequently, effective attack strategies may differ across two fine-tuning modes, motivating separate analyses in our benchmark.

\textbf{Target properties: the demographic information and the medical diagnosis frequency.}
Since two fine-tuned modes have different memorization patterns, property inference behavior can vary depending on where the target property resides.
We therefore propose two categories of the properties: the demographic information, which is often revealed in the patient description, and the medical diagnoses, which are discussed by both the patients and the doctor, as shown in Figure\ref{fig:ChatDoctor dataset}.

For demographic information, we select patient \textit{gender}, which can be explicitly stated (e.g., "I am female") or implicitly suggested (e.g., "pregnancy" or "periods") in patient descriptions $x$.
We label the gender property using ChatGPT-4o and filter out samples with ambiguous gender indications.
This results in a gender-labeled dataset of 29,791 conversations, in which 19,206 samples have female labels and 10,585 have male labels.
We use 15,000 samples to train the target models and the remaining 14,791 as auxiliary data for evaluating attacks in the grey-box setting.
For medical diagnosis attributes, we use the original training split of the ChatDoctor dataset with size $50,000$ for training the target models and consider three binary properties:
(1) Mental disorders (5.10\%),
(2) Digestive disorders (12.68\%), and
(3) Childbirth (10.6\%).
Please see Appendix~\ref{subsec:labeling} for details on the labeling process and Section~\ref{subsec:expattacksetup} for details on task definitions and model fine-tuning procedures.




\section{Attacks} \label{sec: attack}

Recall that the goal of the adversary is to estimate the value of the property of interest for the target model $\TM$. Prior work \citep{zhou2021propertyinferenceattacksgans, suri2022formalizingestimatingdistributioninference} has given attacks that can achieve this goal on simpler models or image and tabular data, and therefore these do not apply directly to the LLM setting. 
Inspired by the initial ideas from the old attacks, we proposes two new attacks tailored for the LLM setting.



\subsection{Generation-Based Attack under Black-Box Setting}\label{subsec:genattack}
Prior work \cite{zhou2021propertyinferenceattacksgans} introduced an output-generation-based property inference attack under black-box access, specifically targeting unconditional GANs. In the context of LLMs, which perform conditional token-level generation, we adapt this approach by generating outputs based on carefully designed input prompts that constrain the generation distribution to the domain of interest. Our adapted attack consists of the following three steps.

\textbf{Prompt-conditioned generations.} 
We construct a list of prompts $T$, that encodes high-level contextual information about the fine-tuning dataset.
For example, for the Chat-Doctor dataset, a prompt like ``\textit{Hi, doctor, I have a medical question.}" would be a reasonable choice.
Given any prompt $t\in T$, we generate a corresponding set of output samples $S_{f, t}$ from the target model $f$.

\textbf{Property labeling.} We define a property function $\hat{P}$ hold by the adversary, which maps each generated sample $s\in S_{f, t}$ to a value in $\{0, 1, N/A\}$. 
A label of $1$ or $0$ indicates whether the sample reflects the presence or absence of the target property respectively.
$\hat{P}$ assigns the label $N/A$ for the samples that are ambiguous or indeterminate with respect to the property.

\textbf{Prompt-based property inference.} 
To estimate the property ratio, we first restrict attention to samples with valid labels.
Let $S_{f, t}^*\subseteq S_{f, t}$ denote the subset of generated samples for which $\hat{P}(s)\neq N/A$.
The estimated ratio given the prompt $t$ is
$
\hat{r}_{t} = \frac{1}{|S_{f, t}^*|}\sum_{s\in S_{f, t}^*}\hat{P}(s).
$
If the adversary uses a list of prompts $T$, the aggregated estimation across prompts is given by:
$
$
$\hat{r} = \frac{1}{|T|}\sum_{t\in T} \hat{r}_{t}$.

\subsection{Shadow-Model Attack with Word Frequency under Grey-Box Setting} \label{shadow_attack}


Prior work~\citep{suri2022formalizingestimatingdistributioninference, suri2024dissectingdistributioninference, hartmann2023distribution} has proposed various shadow-model based property inference attacks. 
The core idea is that the adversary trains multiple shadow models on an auxiliary dataset that is disjoint from the target model's dataset, with varying target property ratios. 
Given both the shadow models and their ground-truth property ratios, the adversary can learn a mapping from some extracted model features to the underlying property ratios. The framework \footnote{Prior work frames property inference as a hypothesis testing problem between two candidate ratios. Our framework extends the existing framework by enabling the adversary to predict property ratios directly.
} 
is describes as follows: 
\begin{enumerate}[leftmargin=*, nosep]
\item \textbf{Shadow model training.} The adversary selects $k_1$ target property ratios $r_1, \ldots, r_{k_1} \in [0, 1]$. For each ratio $r_i$, the adversary subsamples $k_2$ auxiliary datasets to match $r_i$ with the target size $n$, and fine-tunes LLMs with the same fine-tuning procedure $\mathcal{A}$, resulting in $k_1 \cdot k_2$ shadow models. 
The shadow models can be denoted as $f_{i, j}$, where $i$ indexes the ratio, and $j$ indexes the repetition.
\item\textbf{Meta attack model training through a defined shodow feature function.} A \textit{shadow feature function} $F$ maps each model to a $d$-dimensional feature vector. Given the shadow models and their corresponding ratios, a meta dataset is constructed: ${(F(f_{i, j}), r_i) \mid i \in [k_1], j \in [k_2]}$. A meta attack model $g: \mathbb{R}^d \to [0, 1]$ is learned from the meta dataset to predict the property ratio from the extracted model features. In this paper, we use XGBoost~\citep{chen2015xgboost} to train the meta attack model.
\item \textbf{Property inference.} The final inference on the target model $f$ is made by computing $\hat{r}=g(F(f))$.
\end{enumerate}

\textbf{Constructing new shadow attacks with word frequency.}
The choice of the shadow feature functions $F$ plays an important role in the success of the attack.
While previous work relies on loss or probability vector \cite{suri2022formalizingestimatingdistributioninference, suri2024dissectingdistributioninference} , some studies have shown that these features may not be the most effective way to measure the performance of the LLMs \cite{carlini2021extractingtrainingdatalarge, duan2024membershipinferenceattackswork}. 
We alternatively focus on another feature specific to the LLM setting, \textbf{\textit{word frequency}}, which has been parallelly studied for the literature of membership inference attack \cite{kaneko2024samplingbasedpseudolikelihoodmembershipinference, meeus2025canarysechoauditingprivacy}.
Our attack is based on the intuition that certain properties may strongly correlate with the appearance of specific words in the text.
As a result, models fine-tuned on datasets with different property distributions may exhibit distinct word patterns in their generations.

Assume $V^*$ is a selected list of $d$ keywords, which we will describe its construction later.
Similar to the generation attack, given a model $f$ and the prompt $t$ that describes the meta information about the fine-tuning dataset, we generate a set of text samples $S_{f,t}$.
For each word $v\in V^*$, we calculate the word-frequency $\mu_v^{f,t}$, defined as the proportion of samples in $S_{f,t}$ containing $v$. If the adversary uses a list of prompts, it can average this by $u_{v}^f = \frac{1}{T}\sum_{t \in T} u_{v}^{f,t}$.
The resulting vector $(\mu_v^f)_{v\in V^*}\in [0, 1]^d$ serves as the shadow feature, and the shadow feature function is defined as 
$F_{\rm word}(f):=(\mu_v^f)_{v\in V^*}.$

To construct the keyword list $V^*$, 
we first define the full vocabulary $V$ as all words that appear in at least one sample in any $S_{f_{i, j},t}$.
Then we apply a standard feature selection algorithm \footnote{We used the algorithm \texttt{f\_regression} implemented in scikit-learn library~\citep{pedregosa2011scikit}.} using the word frequency $(\mu_v^{f_{i,j}})_{v\in V}$ and their corresponding labels(i.e. the property ratios).
This process selects the $d$ most informative words for the property ratio prediction task, forming the final keyword list $V^*$.


\section{Experiments}

In this section, we empirically evaluate the effectiveness of our proposed attacks within the newly introduced benchmark, \taskname.
Specifically, we aim to answer the following research questions:
\begin{enumerate}[leftmargin=*,nosep]
\item How do the proposed attacks perform in Chat-Completion Mode versus Q\&A Mode?
\item How does the choice of fine-tuning method influence the success of property inference attacks?
\end{enumerate}

\subsection{Experimental Setup}\label{subsec:expattacksetup}
For implementation details, including the selection of hyperparameters for fine-tuning, our attacks, and baseline methods, please refer to Appendix~\ref{subsec:experiment setup}.

\textbf{Models.}~We use three open base models for experimentation: Llama-1-8b\citep{touvron2023llama}, Pythia-v0-6.9b\cite{biderman2023pythia} and Llama-3-8b-instruct \citep{llama3modelcard}. We use the Llama-1 and Pythia-v0 since these were released before the original ChatDoctor dataset and hence have no data-contamination from the pre-training stage, giving us a plausibly more reliable attack performance. While Llama-3 came after ChatDoctor release, we still use it since it is highly performant and is widely used for experimentation. Refer to Appendix \ref{subsec:experiment setup} for implementation details and fine-tuning performance.

\textbf{Property inference tasks.} Our benchmark defines two property inference tasks.
\textbf{Gender property inference}, where the goal is to infer the ratios of female samples in the fine-tuning dataset. 
We define $3$ target ratios of female: $\{0.3, 0.5,0.7 \}$; for each target ratio, we subsample 3 datasets with different random seeds to match each target ratio while keeping the same size $6500$, and we evaluate this by attacking the total 9 target models.
\textbf{Medical diagnosis property inference}, where the goal is to infer the proportion of three diagnosis-related properties (e.g., mental disorder $(5.10\%)$, digestive disorder($12.68\%$), childbirth($10.6\%$) from the medical diagnosis dataset(with size $50,000$). We train $3$ target models on the entire dataset for evaluation.

For both tasks, we evaluate the attacks on Q\& A Mode and Chat-Completion Mode.
For the gender inference task, we evaluate both black-box and grey-box attacks, where our benchmark provides auxilary dataset of size $14,791$. 
For the medical diagnosis task, we evaluate only the black-box adversary, as the grey-box setting requires that the auxiliary dataset shares the same conditional distribution given the target property while differing only in the marginal distribution. Constructing a well-matched auxiliary dataset for multiple properties simultaneously is inherently nontrivial.

\textbf{Our attack setups.} 
For the \textbf{black-box generation-based attack (BB generation)}  as described in Section~\ref{subsec:genattack} on our benchmark, one example of the prompts we used is to fill out the sentence: "Hi, Chatdoctor, I have a medical question." 
In total, we use three prompts; the full list is provided in Appendix~\ref{subsec:experiment setup}. 
For each target model $f$ and prompt $t$, we generate $2000$ samples. 
Each generated text is then labeled by ChatGPT-4o ($\hat{P}$) based on the target property.

For the \textbf{shadow-model attack with word frequency (word-frequency attack)}, as described in Section~\ref{shadow_attack}, we choose $k_1=7$ property ratios in $\{0.2, 0.3, \cdots, 0.8 \}$, with $k_2=5$ or $6$ (varying between different LLMs) shadow models trained per ratio.
We apply the same three prompts as in the BB generation and generate $\sim 100$k samples for each prompt to estimate the word frequency.


\textbf{Baseline attacks.} 
We consider three baseline attacks and put some implementation details in Appendix \ref{subsec:experiment setup}. (1) \textbf{Direct asking} (black-box baseline) is a direct query approach, where the adversary simply asks the model to report the property ratio.
For example, we prompt the model with: "what is the percentage of patient having mental disorder concern in the ChatDoctor dataset?".
(2) \textbf{Perplexity attack} (grey-box baseline) is the shadow-model attack leveraging perplexity score as the shadow features instead of word-frequency. We keep the remaining set-ups the same as our word-frequency attack.
(3) \textbf{Generation w/o FT} (sanity-check baseline) is the generation-based attack on \emph{pretrained LLMs}, which helps ensure that the success of our method is not simply due to prior knowledge encoded during pretraining. We evaluate this baseline for three medical diagnosis properties, but exclude it for the gender attribute, since our evaluations already involve varying gender ratios.

\textbf{Attack Evaluation.} Since the adversary aims to infer the exact property ratio, which is a continuous number between 0 and 1, we follow \cite{zhou2021propertyinferenceattacksgans} and use the absolute error between predicted ratio $\hat{r}$ and groundtruth property ratio $r$ to evaluate the attack performance, defined by $|r - \hat{r}|$. The adversary is said to perfectly estimate the target ratio when the absolute error is zero. 


\subsection{Main Results}
\begin{table*}[t]
\caption{\textbf{Attack Performance for gender property in the Q\&A mode and Chat-Completion mode.} Reported numbers are the Mean Absolute Errors (MAE; $\downarrow$) between the predicted and target ratios. 
We highlight the attack that achieves the smallest total MAE across different target ratios.}
\centering
\footnotesize
\resizebox{\textwidth}{!}{
\begin{tabular}{ll|ccc|ccc}
\toprule
\textbf{Model} & \textbf{Attacks} & \multicolumn{3}{c|}{\textbf{Q\&A Mode}} & \multicolumn{3}{c}{\textbf{Chat-Completion Mode}} \\
              &                   & 30 & 50 & 70 & 30 & 50 & 70 \\
\midrule
\multirow{4}{*}{Llama-1}
& Direct asking         & $23.17{\scriptstyle \pm 1.78}$ & $3.98{\scriptstyle \pm 1.88}$ & $18.6{\scriptstyle \pm 0}$ & $22.8{\scriptstyle \pm 1.98}$ & $7.7{\scriptstyle \pm 1.8}$ & $18.57{\scriptstyle \pm 4.71}$ \\
& BB generation         & $36.52{\scriptstyle \pm 0.11}$ & $15.45{\scriptstyle \pm 3.09}$ &  $1.45{\scriptstyle \pm 0.64}$ & \cellcolor{gray!20} $1.73{\scriptstyle \pm 0.76}$ & \cellcolor{gray!20} $2.64{\scriptstyle \pm 3.33}$ & \cellcolor{gray!20} $3.28{\scriptstyle \pm 3.64}$ \\
& Perplexity            & $28.67{\scriptstyle \pm 9.34}$ & $9.38{\scriptstyle \pm 8.95}$ & $24.16{\scriptstyle \pm 2.45}$ & $35.19{\scriptstyle \pm 10.99}$ & $14.5{\scriptstyle \pm 5.98}$ & $5.33{\scriptstyle \pm 6.09}$ \\
& Word-frequency        & \cellcolor{gray!20}$11.43{\scriptstyle \pm 3.0}$ & \cellcolor{gray!20}$7.33{\scriptstyle \pm 6.59}$ & \cellcolor{gray!20}$6.85{\scriptstyle \pm 5.03}$ & $3.44{\scriptstyle \pm 4.61}$ & $0{\scriptstyle \pm 0}$ & $6.6{\scriptstyle \pm 9.35}$ \\
\midrule
\multirow{4}{*}{Pythia-v0}
& Direct asking\footnotemark         & -- & -- & -- & -- & -- & -- \\
& BB generation         & $46.75{\scriptstyle \pm 3.64}$ & $23.45{\scriptstyle \pm 5.89}$ & $10.31{\scriptstyle \pm 4.85}$ &\cellcolor{gray!20} $3.56{\scriptstyle \pm 2.03}$ & \cellcolor{gray!20} $ 5.61{\scriptstyle \pm 0.78}$ & \cellcolor{gray!20} $2.15{\scriptstyle \pm 2.45}$ \\
& Perplexity            & $22.33{\scriptstyle \pm 15.8}$ & $11.25{\scriptstyle \pm 13.68}$ & $25.79{\scriptstyle \pm 15.59}$ & $4.32 {\scriptstyle \pm 3.25}$ & $9.94 {\scriptstyle \pm 0.59}$ & $9.39 {\scriptstyle \pm 0}$ \\
& Word-frequency        & \cellcolor{gray!20}$22{\scriptstyle \pm 10.4}$ & \cellcolor{gray!20}$7.95{\scriptstyle \pm 9.44}$ & \cellcolor{gray!20}$9.25{\scriptstyle \pm 11.9}$ & $3.31{\scriptstyle \pm 4.68}$ & $3.27{\scriptstyle \pm 4.62}$ & $6.73{\scriptstyle \pm 8.22}$ \\
\midrule
\multirow{4}{*}{Llama-3}
& Direct asking         & $14.27{\scriptstyle \pm 5.32}$ & $4.86{\scriptstyle \pm 1.33}$ & $19.9{\scriptstyle \pm 4.24}$ & $17.97{\scriptstyle \pm 5.33}$ & $4.0{\scriptstyle \pm 2.12}$ & $16.17{\scriptstyle \pm 1.18}$ \\
& BB generation         & $23.64{\scriptstyle \pm 5.82}$ & $5.79{\scriptstyle \pm 6.46}$ & $14.01{\scriptstyle \pm 1.68}$ & \cellcolor{gray!20}$0.61{\scriptstyle \pm 0.77}$ & \cellcolor{gray!20}$1.33{\scriptstyle \pm 1.31}$ & \cellcolor{gray!20}$1.25{\scriptstyle \pm 1.52}$ \\
& Perplexity            & $13.28{\scriptstyle \pm 4.77}$ & $25.0{\scriptstyle \pm 25.4}$ & $19.01{\scriptstyle \pm 20.52}$ & $17.80{\scriptstyle \pm 9.06}$ & $19.85{\scriptstyle \pm 7.6}$ & $6.24{\scriptstyle \pm 7.57}$ \\
& Word-frequency        & \cellcolor{gray!20}$8.29{\scriptstyle \pm 2.13}$ & \cellcolor{gray!20}$7.33{\scriptstyle \pm 6.59}$ & \cellcolor{gray!20}$10.66{\scriptstyle \pm 7.12}$ & $2.45{\scriptstyle \pm 2.3}$ & $3.33{\scriptstyle \pm 4.7}$ & $5.83{\scriptstyle \pm 1.73}$ \\
\bottomrule
\end{tabular}
}
\label{tab:gender}
\end{table*}
\footnotetext{The fine-tuned Pythia model fails to produce any output when queried with direct prompts, so its performance cannot be meaningfully evaluated. The same issue arises with the pretrained Pythia model, likely due to its limited instruction-following capabilities. \label{ft:pythia}}

\paragraph{Gender property inference.}
Table~\ref{tab:gender} presents the results of our attacks on the gender property inference task for models fine-tuned in both Q\&A Mode and Chat-Completion Mode.
We highlight two main observations:
First, \textbf{in Q\&A Mode, our word-frequency attack significantly outperforms both baselines and our BB generation attack.}
Second, \textbf{in Chat-Completion Mode, the BB generation attack achieves the best performance, with the word-frequency attack performing closely behind -- both substantially outperforming the baselines}.

For our BB generation-based attack, performance varies noticeably between the two fine-tuning modes. For example, for the Llama-1 model, the Q\&A mode yields a high MAE of $17.5\%$, whereas the CC mode achieves a much lower MAE of $2.55\%$.
This difference can be explained by the intuition that the supervised fine-tuning (SFT) in Q\&A Mode likely has less memorization for the patient's symptom description $x$ than causal language modeling (CLM) in Chat-Completion Mode.
Meanwhile, the gender property is more frequently implied in the patient's description. Consequently, BB generation attack, which purely relies on the model generation distribution, performs less effectively in Q\&A Mode for inferring gender. 

Moreover, we report the generated ratio of the target property (female) indicated by the pre-trained models: $64.2\%$ (Llama-1),	$67.7\%$ (Pythia),	$62.6\%$ (Llama-3). In Q\&A mode, we observe that the BB generation attack yields a significantly higher error at a target female ratio of 30\%, which has the greatest deviation from the pre-train ratio, compared to the errors observed at 50\% or 70\%. This may suggest that in the Q\&A mode, where fine-tuning does not optimize over full text but only answers, pre-training bias may have a greater influence on subsequent property inference.

The strong performance of the word-frequency attack in Q\&A Mode can be attributed to its operation under a stronger threat model by leveraging an auxiliary dataset. Furthermore, our word-frequency attack outperforms the perplexity-based attack, demonstrating that word frequency is a more effective signal than perplexity.
Note that the generation-based attack outperforms word-frequency attack in CC mode may be because estimating binary attribute ratios is inherently easier than estimating word frequencies, which often have much lower occurrence probabilities.

\paragraph{Medical diagnosis property inference.}
Table~\ref{tab:medical} presents the results of our attacks on the medical diagnosis property inference task for models fine-tuned in both Q\&A Mode and Chat-Completion Mode.
We highlight two main observations that are consistent across both fine-tuning modes and all three LLMs:
First, \textbf{our BB generation attack achieves strong performance and consistently outperforms both baselines across all three diagnosis attributes.}
Second, \textbf{the attack performs relatively worse on the childbirth attribute compared to mental disorder and digestive disorder.}

Interestingly, 
unlike the gender property task, the BB generation attack achieves strong performance in two both modes, we suspect the reason is that the medical diagnosis properties are strongly reflected in both the patient input and the doctor’s response (e.g. Figure~\ref{fig:ChatDoctor dataset}).


The relatively lower performance on the childbirth attribute may be explained by the results of the Generation w/o FT baseline. For example, the exact ratios of three medical properties from the pre-trained Llama-3 model are $1.65\%$ (mental disorder), $8.03\%$ (digestive disorder) and $0.279\%$ (childbirth). We observe that the pre-train ratio for childbirth is notably lower than the pre-train ratio for the other two properties.
We suspect this is due to the cultural sensitivity of childbirth-related topics (e.g., pregnancy, abortion), which may have led to safety training during pretraining that suppresses the generation of such content.
As a result, the pretrained model’s output distribution is likely the most misaligned with the fine-tuned target distribution for this property, reflected by the highest MAE among the three attributes.
This might limits the effectiveness of our attack.

\begin{table*}[t]
\centering
\footnotesize
\vspace{-3ex}
\caption{\textbf{Attack Performance for medical diagnosis in the Q\&A mode and Chat-Completion mode.} Reported numbers are the Mean Absolute Errors (MAE; $\downarrow$) between the predicted and target ratios. 
We highlight the attack that achieves the smallest total MAE across different target properties.}
\resizebox{\textwidth}{!}{
\begin{tabular}{ll|ccc|ccc}
\toprule
\textbf{Model} & \textbf{Attacks} & \multicolumn{3}{c|}{\textbf{Q\&A Mode}} & \multicolumn{3}{c}{\textbf{Chat-Completion Mode}} \\
              &                   & Mental & Digestive & Childbirth & Mental & Digestive & Childbirth \\
\midrule
\multirow{3}{*}{Llama-1}
& Generation w/o FT     & 3.45 & 4.19 & 9.88 & 3.45 & 4.19 & 9.88 \\
& Direct asking         & $7.66{\scriptstyle \pm 2.05}$ & $0.18{\scriptstyle \pm 0}$ & $9.2{\scriptstyle \pm 0}$ & $8.62{\scriptstyle \pm 2.36}$ & $0.17{\scriptstyle \pm 0}$ & $9.2{\scriptstyle \pm 0}$ \\
& BB generation         & \cellcolor{gray!20}$2.55{\scriptstyle \pm 0.25}$ & \cellcolor{gray!20}$3.94{\scriptstyle \pm 0.37}$ & \cellcolor{gray!20}$7.95{\scriptstyle \pm 0.36}$ & \cellcolor{gray!20}$1.76{\scriptstyle \pm 0.23}$ & \cellcolor{gray!20}$1.44{\scriptstyle \pm 0.24}$ & \cellcolor{gray!20}$6.99{\scriptstyle \pm 0.18}$ \\
\midrule
\multirow{3}{*}{Pythia-v0}
& Generation w/o FT     & 1.84 & 9.57 & 9.85 & 1.84 & 9.57 & 9.85 \\
& Direct asking\footref{ft:pythia}         & -- & -- & -- & -- & -- & -- \\
& BB generation         & \cellcolor{gray!20}$1.82{\scriptstyle \pm 0.56}$ & \cellcolor{gray!20}$3.71{\scriptstyle \pm 0.82}$ & \cellcolor{gray!20}$7.63{\scriptstyle \pm 0.31}$ & \cellcolor{gray!20}$1.88{\scriptstyle \pm 0.36}$ & \cellcolor{gray!20}$1.84{\scriptstyle \pm 0.16}$ & \cellcolor{gray!20}$6.23{\scriptstyle \pm 0.52}$ \\
\midrule
\multirow{3}{*}{Llama-3}
& Generation w/o FT     & 3.45 & 4.64 & 10.32 & 3.45 & 4.64 & 10.32 \\
& Direct asking         & $19.96{\scriptstyle \pm 17.74}$ & $14.22{\scriptstyle \pm 0}$ & $10.26{\scriptstyle \pm 0.47}$ & $5.03{\scriptstyle \pm 0}$ & $12.64{\scriptstyle \pm 0}$ & $10.27{\scriptstyle \pm 0.5}$ \\
& BB generation         & \cellcolor{gray!20}$1.43{\scriptstyle \pm 0.7}$ & \cellcolor{gray!20}$1.80{\scriptstyle \pm 1.18}$ & \cellcolor{gray!20}$7.73{\scriptstyle \pm 0.38}$ & \cellcolor{gray!20}$0.63{\scriptstyle \pm 0.23}$ & \cellcolor{gray!20}$1.82{\scriptstyle \pm 0.45}$ & \cellcolor{gray!20}$4.59{\scriptstyle \pm 0.35}$ \\
\bottomrule
\end{tabular}}

\label{tab:medical}
\end{table*}

\textbf{Takeaway.} Our results show that the shadow-model attack with word frequency is particularly effective when the target model is fine-tuned in the Q\&A Mode and the target property is more explicitly revealed in the question than in the answer.
In contrast, when the model is fine-tuned in Chat-Completion Mode or when the target attribute is embedded with both question and answer, the generation-based attack proves to be simple yet highly effective.



\subsection{Additional studies}
\label{sec:exp_ablation}
\paragraph{Empirical analysis of selected keywords.} To better understand the word-frequency attack, we provide examples of the keyword list chosen by the shadow models in Table \ref{example of keywords}. For each word, we compute the correlation coefficient between its occurrence frequency and the corresponding female ratio. A correlation coefficient close to 0 indicates no correlation, while values near 1 or -1 indicate strong positive or negative correlation. We find that many selected keywords exhibit strong correlations with the target female ratio, supporting the effectiveness of word frequency as a signal. Notably, the correlations are generally stronger when the target model is in CC mode compared to QA mode—aligning with the performance differences observed in Table \ref{tab:gender}, where word-frequency attacks perform better under CC mode than QA mode. More details are provided in Appendix~\ref{subsec:ablation}.

\begin{table}[h!]
\centering
\vspace{-2ex}
\caption{Top 5 chosen Keywords and the corresponding Correlation Coefficients for Llama-1 Models}
\begin{tabular}{@{}lll@{}}
\toprule
\textbf{Model} & \textbf{Keywords (Correlation Coefficients)} \\ 
\midrule
\textbf{Llama-1 (CC mode)} & 
his ($-0.956$), himself ($-0.934$), her ($0.947$), he ($-0.950$), female ($0.960$)\\
\textbf{Llama-1 (QA mode)} & 
\small{female ($0.785$), reluctant ($0.774$), spotting ($0.787$), scanty ($0.821$), ovaries ($0.782$)} \\
\bottomrule
\end{tabular}
\label{example of keywords}
\vspace{-2ex}
\end{table}
\paragraph{Ablation study of our attacks.} We conduct an ablation study to assess the impact of key hyperparameters in both of our proposed attacks.
We include the results of this study for the gender attribute using the LLaMA-3 model.
Additional ablation studies can be found in Appendix~\ref{subsec:ablation}.

\begin{figure}[!t]
\centering
\begin{subfigure}[t]{0.25\textwidth}
    \includegraphics[width=\linewidth]{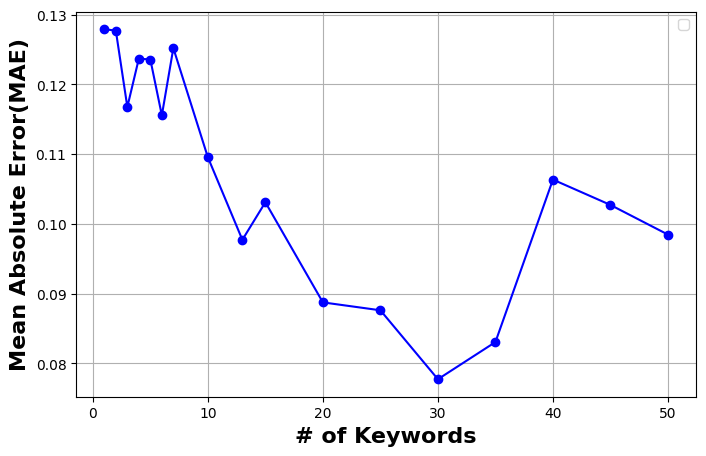}
    \caption{\# of keyword}
    \label{fig:keyword}
\end{subfigure}
\hfill
\begin{subfigure}[t]{0.25\textwidth}
    \includegraphics[width=\linewidth]{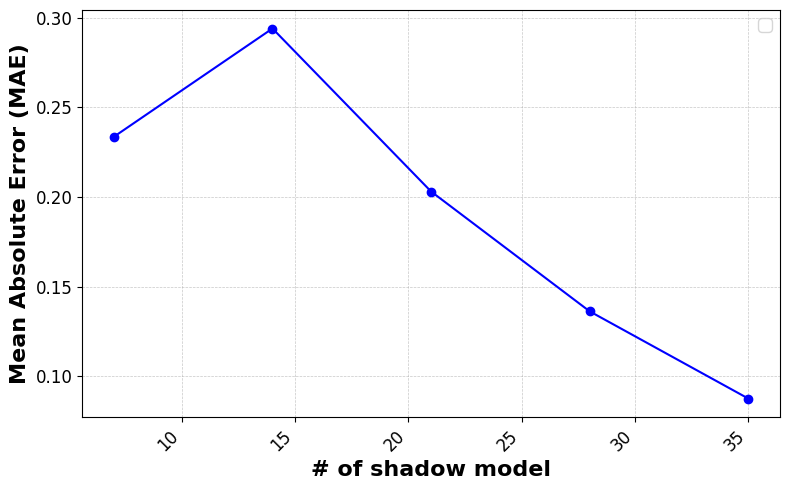}
    \caption{\# of shadow model}
    \label{fig:shadow_model}
\end{subfigure}
\hfill
\begin{subfigure}[t]{0.25\textwidth}
    \includegraphics[width=\linewidth]{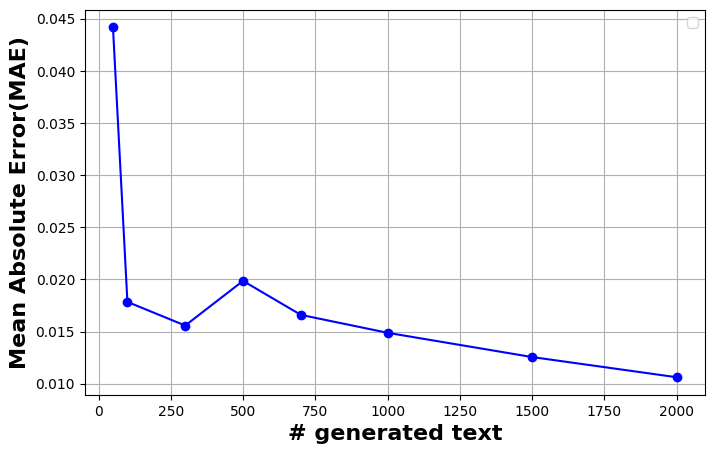}
    \caption{\# generated samples}
    \label{fig:generated_samples}
\end{subfigure}
\caption{Effects of hyperparameters of our attacks for Llama-3 and gender property.}
\vspace{-2ex}
\label{Ablation:Llama3}
\end{figure}

For the BB generation attack, we study how the number of generated samples affects attack performance for the target Chat-Completion Mode model.
As shown in Figure~\ref{fig:generated_samples}, the estimated property ratio converges rapidly: with just 500 samples, the mean absolute error (MAE) drops below $2\%$, indicating the attack's efficiency even under limited query budgets.

For the word-frequency attack, we examine two factors when testing with the Q\&A Mode: the number of selected keywords $d$ and the total number of shadow models $k_1\cdot k_2$.
As shown in Figure \ref{fig:keyword}, the optimal number of keywords lies between 30 and 35.
Using too few keywords may result in weak signals, while too many can introduce noise and overwhelm the meta attack model, given a limited number of shadow models.
Figure~\ref{fig:shadow_model} shows that increasing the number of shadow models improves performance, as it provides more training data for the meta-model, enhancing its generalization.

Additionally, we conduct ablation studies on relaxation on some assumption of the shadow models. To test the assumption of access to the same pre-trained model, we evaluate a setting where the adversary uses Llama-3 for training shadow models but targets a Llama-1 model. MAE results for both Q\& A and CC modes are shown in table \ref{Ablation:model relaxation}. We observe that the knowledge of the pre-trained model has small effects in CC mode, but notable effects in the QA mode. This might be because with the CC mode, the model is learned to fully mimic the texts in the fine-tuning dataset, disregarding what the pre-trained model is. However, in QA mode, while the model learns to answer questions based on the fine-tuning data, the generation of the questions themselves may still rely heavily on the pre-trained model, leading to greater divergence in the resulting question–answer pairs. More experiments can be found in Appendix ~\ref{subsec:ablation}.

\begin{table}[h!]
\centering
\caption{Testing shadow model's assumption on pre-trained model architecture}
\label{tab:combined_performance}
\begin{tabular}{l|ccc|ccc}
\toprule
\multirow{2}{*}{} & \multicolumn{3}{c|}{\textbf{Q\&A Mode}} & \multicolumn{3}{c}{\textbf{Chat-Completion Mode}} \\
\cmidrule{2-7}
& \textbf{30} & \textbf{50} & \textbf{70} & \textbf{30} & \textbf{50} & \textbf{70} \\
\midrule
Knowing the pre-trained model & $6.86\%$ & $12.96\%$ & $6.82\%$ & $3.44\%$ & $0.27\%$ & $6.26\%$ \\
Without knowing the model & $20.23\%$ & $0.24\%$ & $20.00\%$ & $3.85\%$ & $0.17\%$ & $15.31\%$ \\
\midrule

\end{tabular}
\vspace{-3ex}
\label{Ablation:model relaxation}
\end{table}

\section{Discussion}
\label{sec:discuss}

\textbf{Individual privacy vs. dataset-level confidentiality.}
Most prior work on privacy-preserving machine learning looks at individual privacy\cite{carlini2021extractingtrainingdatalarge, shokri2017membership, carlini2022membership}, where the goal is to protect sensitive data information corresponding to each individual. In contrast, our work, as well as the literature on property inference, focuses on the confidentiality of certain aggregate information about a dataset. This kind of confidentiality may be required for several reasons. First, dataset-level properties may reveal strategic business information: a model fine-tuned on a customer-service chat dataset may reveal that the company primarily serves low-income customers, which is some information the company might prefer to keep private. Secondly, dataset-level properties might be sensitive: a hospital with many patients diagnosed with a sensitive condition such as HIV may avoid disclosing this to prevent potential stigma.

\textbf{Possible defenses.} 
Even though it is impossible to provide confidentiality for all properties of a dataset and still produce a useful model, in most practical cases only a small subset of properties are confidential, and these are often largely unrelated to the intended use of the model. For example, the income-level of the customers is unrelated to answering customer service questions.

One plausible defense strategy is to subsample the training data, resulting in a dataset more closely aligned with a known public prior. Although subsampling can mitigate property inference attacks at their source, it may also compromise model utility by limiting the amount of effective training data.
An alternative approach is to adjust the output sampling methods: since the attack depends on model generations, we implement a simple defense that adjust the temperature parameter in the final softmax layer, so that the generation does not reflect the ground-truth distribution; the experiment details are put in Appendix \ref{susec: defense}. However, such a defense may break down if the adversary is aware of the default temperature setting or has access to the model weights. A deeper investigation into the limitations of this defense, as well as the development of more robust mitigation strategies, is an important direction for future work. 

\section{Acknowledge}
This work was supported in part by the ONR under grants N000142412304 and N00014-25-1-2116, by the NSF under grants CIF-2402817 and CNS-2241100, and by the ARO-MURI under grant W911NF2110317.

We thank Yiwei Fu for assistance with checking the GPT-based labeling used in this work.
\section{Conclusion}\label{conclusion}

In conclusion, we introduce a new benchmarking task --\taskname-- for property inference in LLMs and show that property inference can be used to breach confidentiality of fine-tuning datasets; this goes beyond prior work in classification and image generative models. Our work also proposes new property inference attacks tailored to LLMs and shows that unlike simpler models, the precise form of the attack depends on the mode of fine-tuning. We hope that our benchmark and attacks will inspire more work into property inference in LLMs and lead to better defenses. 

\textbf{Limitation and future work.} Firstly, although our attack has a high success rate in inferring the proportion of mental disorder and digestive disorder, it has a low success rate in childbirth; therefore, a natural future work is to propose better attacks to investigate whether there are privacy leakages for childbirth.
Secondly, datasets from other domains can also be relevant under the property inference threat model — for example, capturing the distribution of opinions in a news dataset. Extending the benchmark to include such datasets represents an important direction for strengthening the evaluation.
Lastly, while subsampling can mitigate property inference at its source, it is not ideal when the dataset is limited or the training task requires large amount of data. Hence, more future works on better defenses are needed to protect data confidentiality.

\newpage
\bibliographystyle{abbrv}
\bibliography{sections/ref}

\newpage
\appendix

\section{Technical Appendices and Supplementary Material}


The organization of this appendix is as below:
\begin{itemize}
    \item In Section \ref{subsec:labeling}, we present the details of the labeling process of our benchmark, including the prompts we use to create labels using ChatGPT-4o and the details of the manual inspection.
    \item In Section \ref{subsec:experiment setup}, we present the experimental setup, including fine-tuning details, model performance, and details of the baseline attacks. 
    \item In Section \ref{subsec:ablation}, we present additional ablation study on the impact of the hyperparameters in our proposed attacks. For the word-frequency attack, we study the number of keywords and the number of shadow models over three model architectures and two modes. For the generation attack, we study the number of generated samples and the effect of individual prompts. We also present some selected keywords being used in the main table.
    \item In Section \ref{susec: defense}, we include a simple defense mechanisms.
    
\end{itemize}

\subsection{Labeling process:}\label{subsec:labeling}
\paragraph{Creating gender dataset} We first select samples containing gender related keywords, including "male", "female", "son", "daughter", "pregnancy", "pregnant", "ovary", "man", "woman" etc. Then, we use ChatGPT-4o to create labels with the following prompts:

\textbf{Gender }\textit{You are an assistant that classifies the text based on patient's gender. Is the following text describing the symptoms of a female or male patient? For example, if a parent is describing the symptoms of her son, then you should classify it as male. Please output: 1. female, 2. male, 3. both, 4. unclear: \{ patient's symptom description \} }

Then we filter out ambiguous gender indication and only only select samples with labeling outputs "1. female" or "2. male". The resulting gender dataset has size $29,791$ conversations, in which $19,206$ samples have female labels and $10,585$ samples have male labels.

\paragraph{Creating medical diagnosis dataset} We randomly subsample a subset of the original dataset with size $50,000$. Then we create the labels with the following prompts:

 \textbf{Digestive disorder} \textit{"The following text describes the concerns of a patient and suggestions from a doctor. You are an assistant that classifies the text based on patient's symptoms and doctor's diagnosis.  If the text describes the patient's main concerns about digestive issues, including but are not limited to problems related to stomach, intestine, pancreas, gallbladder or liver,  or describes symptoms related to bloating, diarrhea, constipation, abdominal pain output: digestion. For all other symptoms, output: others: {patient's symptom description}, {doctor's suggestions}}.

 \textbf{Mental disorder} \textit{"The following text describes the concerns of a patient. You are an assistant that classifies the text based on the patient's symptoms. If the text describes a patient's main concern about mental disorder, such as suffering from severe depression, anxiety, or bipolar, output: mental disorder. Note that if the patient simplify express anxiety about other symptoms, or is tired should not be classify as mental disorder.For all other symptoms, output: others: {patient's symptom description}}

 \textbf{Childbirth} \textit{"The following text describes the concerns of a patient. You are an assistant that classifies the text based on the patient's symptoms. If the text describes a patient's main concern about childbirth, preganancy, trying to conceive, or infertility, output: birth. For all the other symptoms, output: others: {patient's symptom description}"}

We only keep ChatGPT outputs with no ambiguous indications. Furthermore, we conduct manual inspections to check the performance of ChatGPT labeling. For the gender dataset, we choose a random subset with size $100$ for manual inspection and $ 100 \%$ of human labeling aligned with the ChatGPT's labeling results. For the medical diagnosis dataset, we choose a random subset with size $200$ for manual inspection; since the context is more complicated and harder for labeling,  $97\%$ of human labeling aligned with ChatGPT's labeling results. 

\subsection{Experiment Setup}\label{subsec:experiment setup}

\paragraph{Experiment compute resources:} All experiments are conducted on NVIDIA RTX 6000 Ada GPU. Each run of the fine-tuning is run on two GPUs; the fine-tuning takes 1.5-3 hours for the smaller fine-tuning dataset (size 6500) and 8-10 hours for the larger fine-tuning dataset (size 50000). Each run of the black-box generation attack is run on 1 GPU. It takes 2-5 hours to generate 100,000 outputs for each model; the time varies on different models.

\textbf{Model fine-tuning details: }Since Llama-1-8b and Pythia-v0-6.9b do not have instruction-following capability, we follow \cite{li2023chatdoctormedicalchatmodel} which first performs instruction fine-tuning on the Alpaca dataset \citep{alpaca}. Next, we fine-tune each model for both QA and chat-completion mode, with supervised fine-tuning and causal language-modeling fine-tuning, where the training objective equation is included in \ref{LLM Preliminary}. We used the LoRA \cite{hu2021loralowrankadaptationlarge} method for fine-tuning with a learning rate of $1e^{-4}$, dropout rate of $0.05$, LoRA rank of $128$ and $5$ epochs.

\paragraph{Target Model performance} As shown in table \ref{tab:gender_eval} and \ref{tab:med_eval}, we evaluate the performance of the target models using the BERT score\cite{zhang2020bertscoreevaluatingtextgeneration}, following \citep{li2023chatdoctormedicalchatmodel}. In particular, we choose a subset with size $500$ from a separate test dataset, iCliniq dataset, provided by \cite{li2023chatdoctormedicalchatmodel}. We generate outputs given the inputs using greedy decoding and calculate the BERT score between the generated texts and the labels. We observe that the fine-tuned Pythia model, as well as the Pythia base model, sometimes outputs an empty string, hence we only calculate the BERT score between non-empty outputs and its corresponding labels. The performance of these models is similar to the performance reported in the paper \cite{li2023chatdoctormedicalchatmodel}.
\begin{table*}[h!]
\centering
\begin{tabular}{llccc}
\hline
\textbf{Dataset} & \textbf{Model} & \textbf{Precision} & \textbf{Recall} & \textbf{F1 Score} \\ 
\hline
\multirow{4}{*}{Gender} & Llama-1 & $0.840 {\scriptstyle \pm 0.003}$ & $0.836{\scriptstyle \pm 0.001}$ &  $0.838 {\scriptstyle \pm 0.002}$ \\
& Llama-3 & $0.823 { \scriptstyle \pm 0.005}$ & $0.837 { \scriptstyle \pm 0.003 }$ &  $0.830 { \scriptstyle \pm 0.004 }$\\
& Pythia & $0.847 { \scriptstyle \pm 0.002 }$ & $0.842 { \scriptstyle \pm 0.001 }$ & $0.844 { \scriptstyle \pm 0.001 }$ \\

\hline
\multirow{4}{*}{Medical Diagnosis} & Llama-1 & $0.843 { \scriptstyle \pm 0.002 }$ & $0.838 { \scriptstyle \pm 0.002 }$ & $0.841 { \scriptstyle \pm 0.002 }$ \\
& Llama-3 & $0.833 { \scriptstyle \pm 0.003 }$ & $0.84 { \scriptstyle \pm 0.002 }$ & $0.836 { \scriptstyle \pm 0.003 }$ \\
& Pythia & $0.8493 { \scriptstyle \pm 0.003 }$ & $0.841 { \scriptstyle \pm 0.0005 }$ & $0.845 { \scriptstyle \pm 0.002 }$ \\
\hline
\end{tabular}
\caption{Target model evaluation using BERT score in Q\&A mode.}
\label{tab:gender_eval}
\end{table*}

\begin{table*}[h!]
\centering
\begin{tabular}{llccc}
\hline
\textbf{Dataset} & \textbf{Model} & \textbf{Precision} & \textbf{Recall} & \textbf{F1 Score} \\ 
\hline
\multirow{4}{*}{Gender} & Llama-1 & $0.8451 { \scriptstyle \pm 0.003}$ & $0.839 {\scriptstyle \pm  0.002}$ & $0.842 { \scriptstyle \pm 0.002 }$\\
& Llama-3 & $0.823 { \scriptstyle \pm 0.004 }$  & $0.838 { \scriptstyle \pm 0.001 }$ & $0.830 { \scriptstyle \pm 0.003  }$  \\

& Pythia & $0.848 { \scriptstyle \pm 0.003 }$ & $0.839 { \scriptstyle \pm 0.002 }$ &  $0.844 { \scriptstyle \pm 0.002 }$\\
\hline
\multirow{4}{*}{Medical Diagnosis} & Llama-1 & $0.847 { \scriptstyle \pm 0.004 }$ & $0.841 { \scriptstyle \pm 0.002 }$ & $0.844 { \scriptstyle \pm 0.003 }$ \\
& Llama-3 & $0.827 { \scriptstyle \pm 0.009 }$ & $0.838 { \scriptstyle \pm 0.004 }$ & $0.832 { \scriptstyle \pm 0.007 }$\\
& Pythia & $0.840{ \scriptstyle \pm 0.009 }$ & $0.8416 { \scriptstyle \pm 0.002 }$ & $0.840 { \scriptstyle \pm 0.006 }$ \\
\hline
\end{tabular}
\caption{Target model evaluation using BERT score in Chat-Completion mode.}
\label{tab:med_eval}
\end{table*}

\paragraph{Prompts for BB generation attack and shadow attack based on word-frequency} We query each target model with three different prompts as follows.

\begin{itemize}
    \item Instruction = "Suppose you use Chat Doctor to consult some medical suggestions, please fill in the sentence. \#\#\# Response: \textbackslash n" 
    \item Prompt 1 = Instruction + "Hi, Chatdoctor, I have a medical question." 
    \item Prompt 2 = Instruction + "Hi, doctor, I have a medical question."
    \item Prompt 3 = Instruction + "Hi Chatdoctor, here is my question."
\end{itemize}
For the BB-generation attack, we label the generated text using the same prompts in Appendix \ref{subsec:labeling}.


\paragraph{Baseline attack: shadow-model attack with perplexity.} 
Following \cite{suri2022formalizingestimatingdistributioninference}, we use the two-dimensional model performance on two hold-out dataset, $S_0$ and $S_1$, where the property ratios are $0\%$ and $100\%$ as the shadow feature function.
This feature captures how well the model performs on data associated with each property value.
The underlying intuition is that models fine-tuned with different property ratios will bring varying performance on data -- a higher proportion of a property may make the model have better performance on data associated with that property.

In the context of LLMs, we adopt perplexity as the performance metric, a widely used measure that reflects how well a language model predicts a given token sequence.
Formally, the perplexity of a model $f$ on a token sequence $t$ is defined as $\text{Perplexity}(f, t):=\exp\left(-\frac{1}{l} \sum_{i=1}^{l} \log f(t_i \mid t_1, t_2, \ldots, t_{i-1})\right)$.
Accordingly, in the baseline method we call \textit{shadow-model attack with perplexity}, the shadow feature function $F_{\rm perp}$ maps each model $f$ to a two-dimensional feature vector representing its average perplexity on:
$
\left(\frac{1}{|S_0|}\sum_{t\in S_0}\text{Perplexity}(f, t), \frac{1}{|S_1|}\sum_{t\in S_1}\text{Perplexity}(f, t)\right).
$

\paragraph{Baseline direct asking prompts: } we use the following prompts to query the target models:
\begin{itemize}
    \item "Please answer the following question: what is the percentage of female patient in the Chat-Doctor dataset? Please only output a fraction between 0 and 1. \#\#\# Response: \textbackslash n The female ratio is "
    \item "Please answer the following question: what is the percentage of patient that has a mental disorder concern in the ChatDoctor dataset? Please only output a fraction between 0 and 1.  \#\#\# Response: \textbackslash n The percentage of mental disorder is "
    \item "Please answer the following question: what is the percentage of patient that has a digestive disorder concern in the ChatDoctor dataset? Please only output a fraction between 0 and 1.  \#\#\# Response: \textbackslash n The percentage of digestive disorder is "
    \item "Please answer the following question: what is the percentage of patient that has a childbirth concern in the ChatDoctor dataset? Please only output a fraction between 0 and 1.  \#\#\# Response: \textbackslash n The percentage of childbirth is "
\end{itemize}

\subsection{Ablation Study}\label{subsec:ablation}


We conduct an ablation study of the following hyperparameters in both of our proposed attacks for the gender property. 

\begin{itemize}
    \item For the word-frequency attack, we study the effect of the number of keywords $d$ and the number of shadow models $ k_1\cdot k_2$ on the attack performance. We also report the selected keyword list for all target models.
    \item For the black-box generation model, we study the effect of individual prompts and the number of generating samples.
\end{itemize}

\begin{figure}[!h]
\centering
\begin{subfigure}[t]{0.3\textwidth}
    \includegraphics[width=\linewidth]{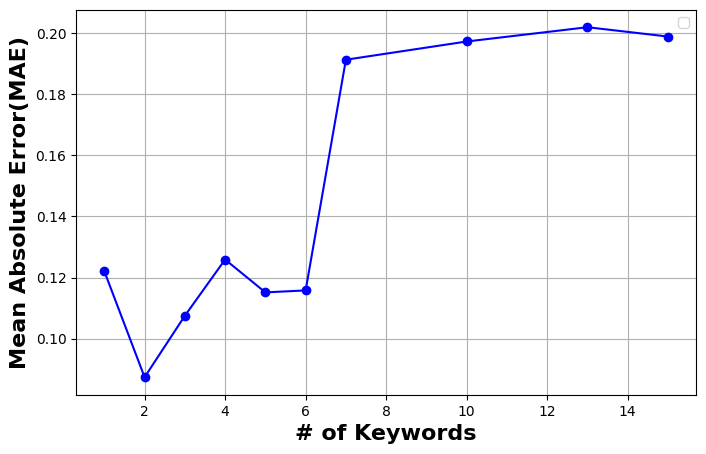}
    \caption{Llama1}
    \label{}
\end{subfigure}
\hfill
\begin{subfigure}[t]{0.3\textwidth}
    \includegraphics[width=\linewidth]{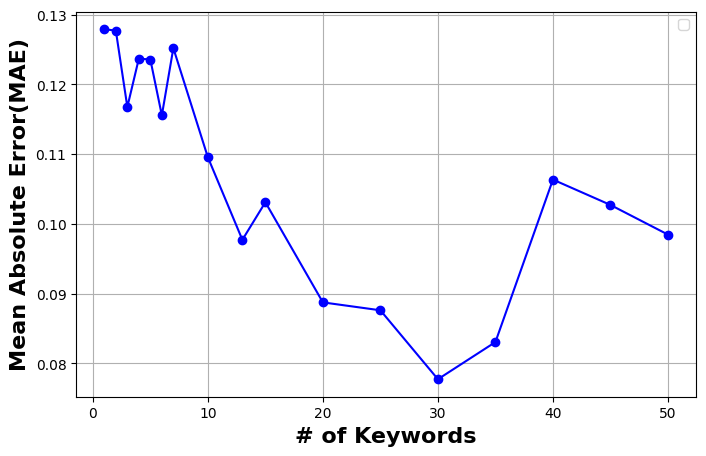}
    \caption{Llama3}
    \label{}
\end{subfigure}
\hfill
\begin{subfigure}[t]{0.3\textwidth}
    \includegraphics[width=\linewidth]{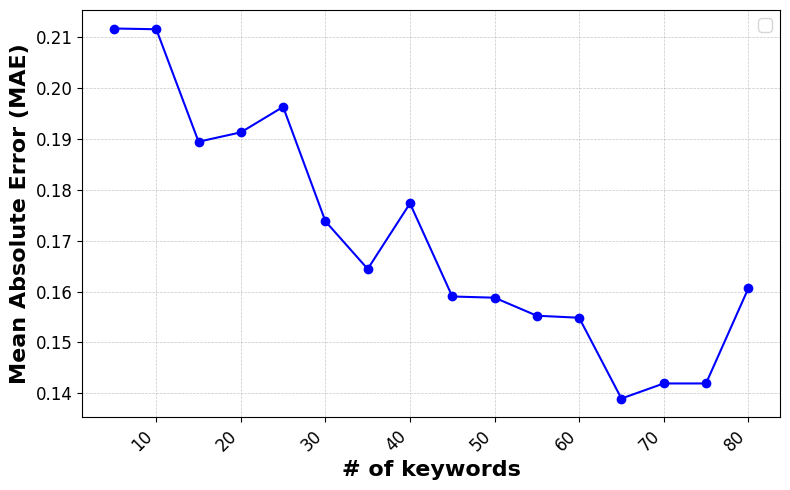}
    \caption{Pythia}
    \label{}
\end{subfigure}
\caption{Effect of number of keywords $d$ for Q\&A mode and gender property. The y axis is the Mean Absolute Errors across different target ratios. }
\label{fig:keyword+QA}
\end{figure}

\paragraph{Ablation study for the Word-frequency attack}
Figure \ref{fig:keyword+QA} shows the ablation study in the Q\&A mode; the optimal number of keywords for word frequency attack varies between different architectures. For the Llama1 model, the optimal number of keywords is less than 5; for example, when $d = 3$, the chosen keywords are "spotting", "female" and "scanty", where "spotting" and "female" are gender-indicated words. 
For the Llama3 model, the optimal number of keywords lies between 30 and 35.
For the pythia model, the optimal number of keywords lies between 65 and 75. We observe that the chosen keywords as well as the number of keywords are very distinct between models; we suspect the reason is that the pre-training data distribution and the model architecture is different for three base models, hence it may have an effect of the generated text distributions. 


\begin{figure}[!h]
\centering
\begin{subfigure}[t]{0.3\textwidth}
    \includegraphics[width=\linewidth]{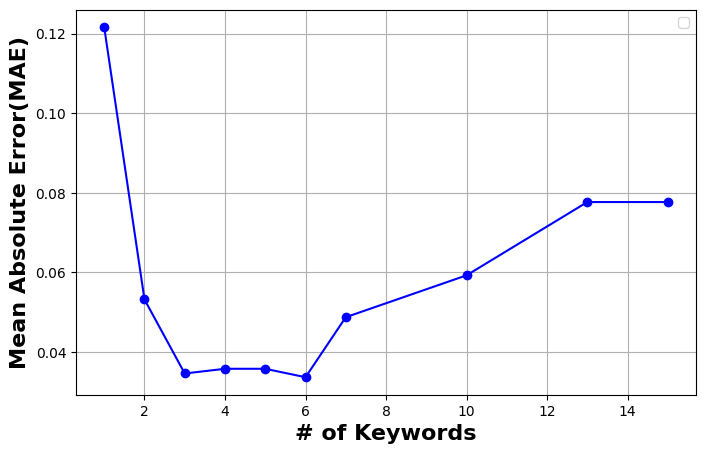}
    \caption{Llama1}
    \label{}
\end{subfigure}
\hfill
\begin{subfigure}[t]{0.3\textwidth}
    \includegraphics[width=\linewidth]{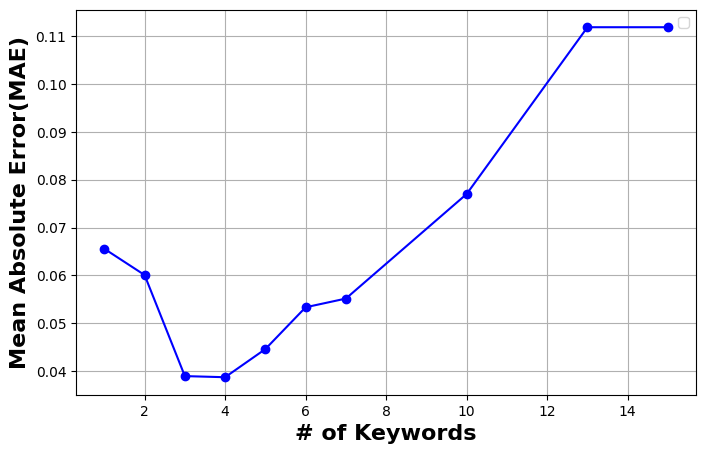}
    \caption{Llama3}
    \label{}
\end{subfigure}
\hfill
\begin{subfigure}[t]{0.3\textwidth}
    \includegraphics[width=\linewidth]{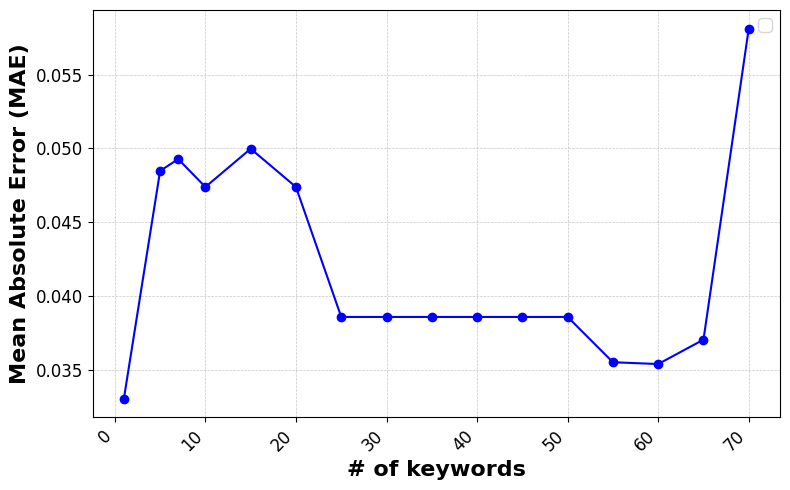}
    \caption{Pythia}
    \label{keyword+CC+Pythia}
\end{subfigure}
\caption{Effect of number of keywords $d$ in Chat-completion mode and gender property. The y axis is the Mean Absolute Errors across different target ratios. }
\label{fig:keyword+CC}
\end{figure}

Figure \ref{fig:keyword+CC} shows the ablation study in the Chat-Completion mode. For Llama1 model, the optimal number of keywords is between $3-6$; when $d = 5$, the chosen keywords are "his", "her", "he", "female", and "she", where all chosen keywords are strongly gender-indicated. For the Llama3 model, the optimal keywords are between $3-5$; when $d = 5$, the chosen keywords are "penile, "female", "scrotal", "masturbating" and "erection", where all chosen keywords are gender-indicated. For the Pythia model, the mean absolute error is less than $5\%$ for $d <70$, which shows that the attack performance is effective; when $d = 5$, the chosen keywords are "scrotum", "penis", "foreskin", "glans" and "female". We observe that in the Chat-Completion mode, all the selected keywords are strongly gender-indicated and with a very small number of keywords, the word-frequency based shadow model attack achieves an effective performance.

Additionally, We put the keyword list chosen by the shadow models in Table \ref{Keyword chosen (complete)}. For each word, we compute the
correlation coefficient between its occurrence frequency and the corresponding female ratio. A
correlation coefficient close to 0 indicates no correlation, while values near 1 or -1 indicate strong
positive or negative correlation, respectively. Notably, for all models, we find that the correlations are generally stronger when the target model is in CC mode compared to QA mode—aligning with the performance differences observed in Table \ref{tab:gender}, where word-frequency attacks perform better under CC mode than QA mode.

\begin{table}[h!]
\centering
\caption{Top Keywords and Correlation Coefficients for Different Models and Modes}
\begin{tabular}{@{}lll@{}}
\toprule
\textbf{Model (Mode)}  & \textbf{Keywords (Correlation Coefficients)} \\ 
\midrule

Llama-1 (CC) &
his ($-0.956$), himself ($-0.934$), her ($0.947$), he ($-0.950$), female ($0.960$), \\
& him ($-0.933$), prostate ($-0.931$), she ($0.954$), son ($-0.929$), daughter ($0.932$) \\[1em]

Pythia (CC) &
scrotum ($-0.941$), he ($-0.916$), penis ($-0.944$), foreskin ($-0.934$), \\ &  male ($-0.928$), glans ($-0.940$), female ($0.968$), masturbate ($-0.919$),  \\
& masturbation ($-0.921$), tip ($-0.917$) \\[1em]

Llama-3 (CC) &
penile ($-0.940$), erect ($-0.935$), penis ($-0.939$), scrotum ($-0.935$), \\ female ($0.980$), 
& scrotal ($-0.941$), males ($-0.935$), masturbating ($-0.945$), \\
& erection ($-0.945$), erectile ($-0.934$) \\[0.3em]
\bottomrule \\
[0.3em]

Llama-1 (QA) &
female ($0.785$), reluctant ($0.774$), spotting ($0.787$), scanty ($0.821$), ovaries ($0.782$) \\ [1em]

Pythia (QA) &
pelvic ($0.738$), recurring ($0.730$), football ($-0.740$), bland ($0.760$), \\ & indigestion ($0.740$), bothering ($0.822$), uti ($0.746$), point ($0.743$),\\
& presenting ($-0.751$), smear ($0.753$) \\[1em]

Llama-3 (QA) &
nifedipine ($-0.786$), readings ($-0.788$), yielding ($-0.765$), squats ($-0.788$), \\ & analogs ($-0.791$),  smoke ($-0.773$), particular ($0.847$), cigarette ($-0.774$), \\
& quit ($-0.810$), regularly ($-0.832$) \\

\bottomrule
\end{tabular}
\label{Keyword chosen (complete)}
\end{table}



In general, we observe that using too few keywords may result in weak signals, while too many can introduce noise and overwhelm the meta-attack models, given a limited number of shadow models. Hence, the optimal $d$ should be in the middle. For Figure \ref{keyword+CC+Pythia}, the MAE of the Pythia model is low ($< 5\%$) for $d<70$; we suspect the reason is that the selected keywords are strongly correlated with gender. 

\begin{figure}[!h]
\centering
\begin{subfigure}[t]{0.3\textwidth}
    \includegraphics[width=\linewidth]{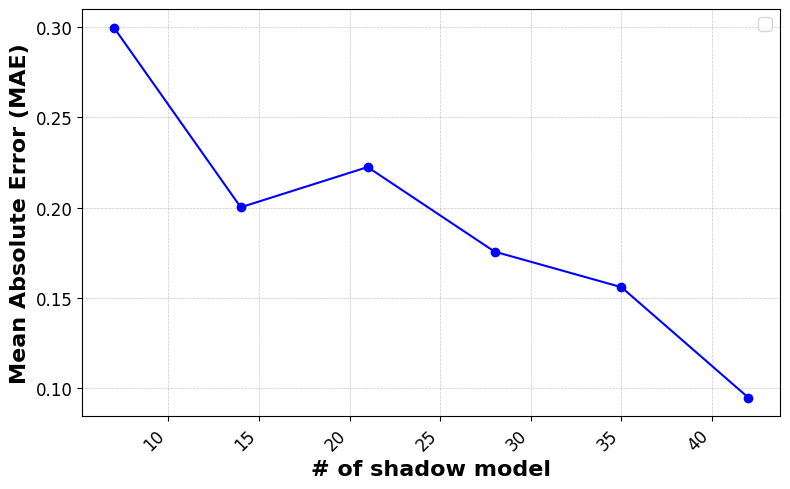}
    \caption{Llama1(d=2)}
    \label{}
\end{subfigure}
\hfill
\begin{subfigure}[t]{0.3\textwidth}
    \includegraphics[width=\linewidth]{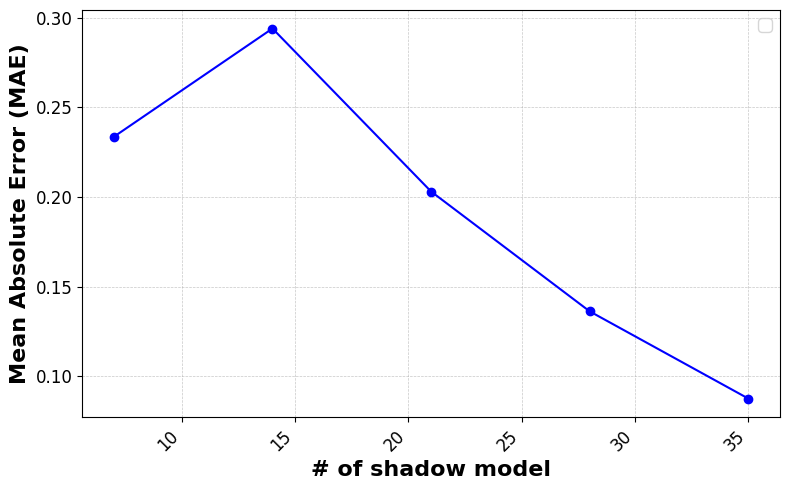}
    \caption{Llama3(d=25)}
    \label{}
\end{subfigure}
\hfill
\begin{subfigure}[t]{0.3\textwidth}
    \includegraphics[width=\linewidth]{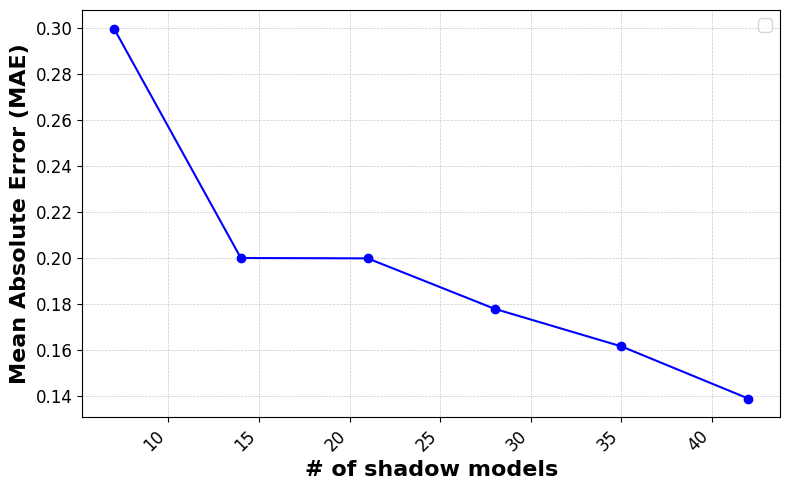}
    \caption{Pythia(d=65)}
    \label{}
\end{subfigure}
\caption{Effect of number of shadow models $k_1 \cdot k_2$ in Q\&A mode and gender property. The y axis is the Mean Absolute Errors across different target ratios. }
\label{fig:shadow_model+QA}
\end{figure}

\begin{figure}[!h]
\centering
\begin{subfigure}[t]{0.3\textwidth}
    \includegraphics[width=\linewidth]{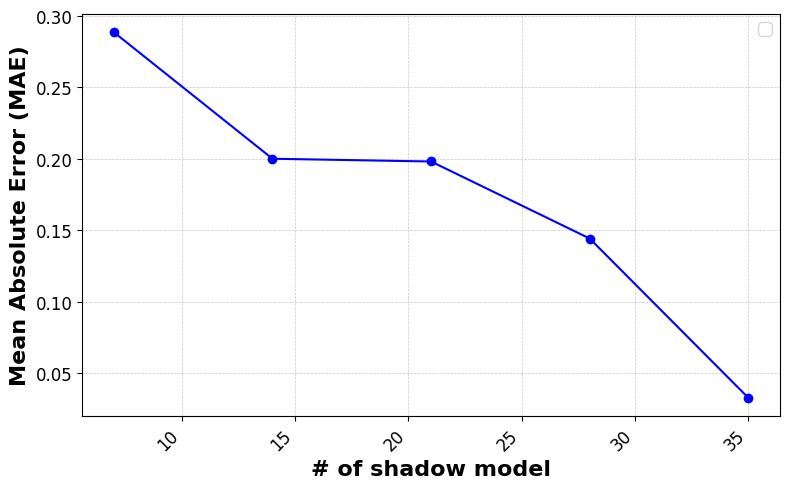}
    \caption{Llama1(d=3)}
    \label{}
\end{subfigure}
\hfill
\begin{subfigure}[t]{0.3\textwidth}
    \includegraphics[width=\linewidth]{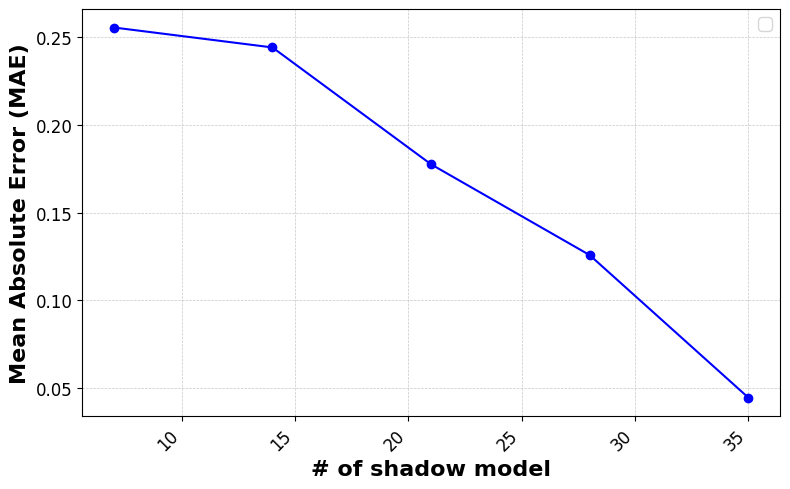}
    \caption{Llama3(d=3)}
    \label{}
\end{subfigure}
\hfill
\begin{subfigure}[t]{0.3\textwidth}
    \includegraphics[width=\linewidth]{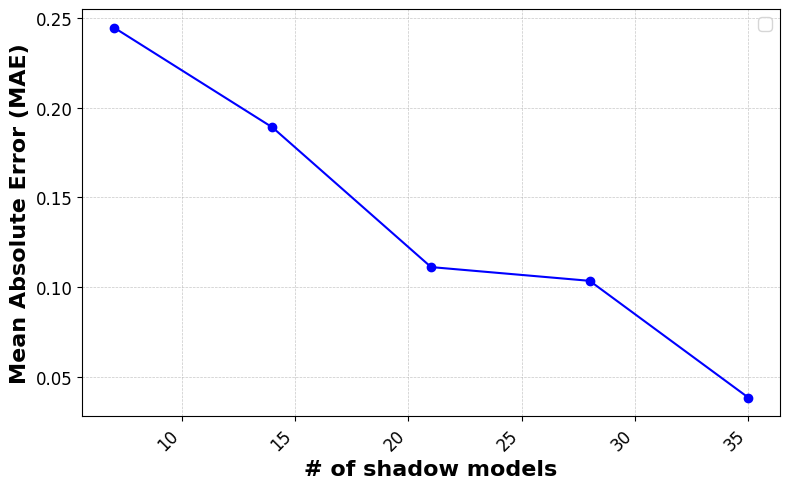}
    \caption{Pythia(d=30)}
    \label{}
\end{subfigure}
\caption{Effect of number of shadow models $k_1 \cdot k_2$ in Chat-Completion mode and gender property. The y axis is the Mean Absolute Errors across different target ratios. }
\label{fig:shadow_model+CC}
\end{figure}

Figure \ref{fig:shadow_model+QA} and \ref{fig:shadow_model+CC} show the effect of the number of shadow models in both the Q\&A mode and the Chat-Completion mode. The figures show that increasing the number of shadow models improves the attack performance, as it provides more training data for the meta-model, enhancing its generalization. 

\begin{figure}[!h]
\centering
\begin{subfigure}[t]{0.3\textwidth}
    \includegraphics[width=\linewidth]{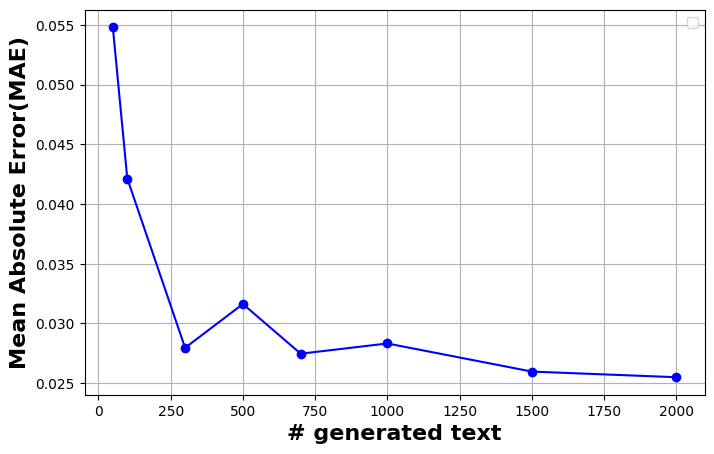}
    \caption{Llama1}
    \label{}
\end{subfigure}
\hfill
\begin{subfigure}[t]{0.3\textwidth}
    \includegraphics[width=\linewidth]{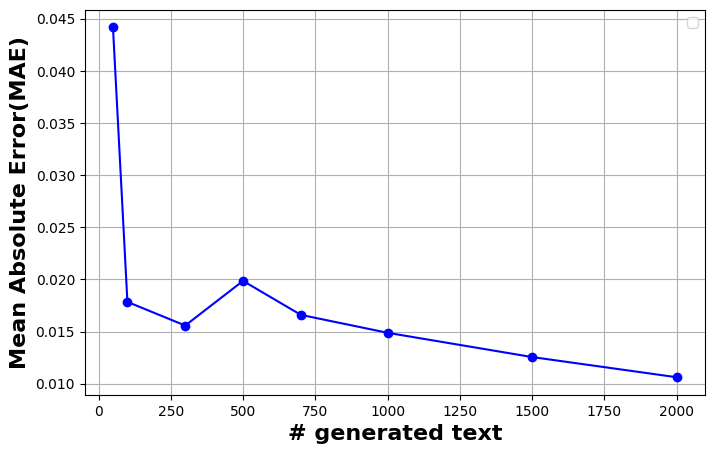}
    \caption{Llama3}
    \label{}
\end{subfigure}
\hfill
\begin{subfigure}[t]{0.3\textwidth}
    \includegraphics[width=\linewidth]{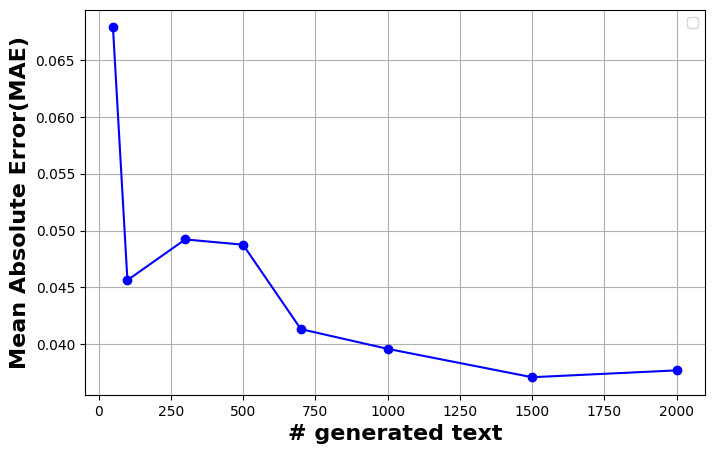}
    \caption{Pythia}
    \label{}
\end{subfigure}

\caption{Effect of number of generated text in Chat-Completion mode for gender property. The y axis is the Mean Absolute Errors across different target ratios. }
\label{fig:samples+CC gender}
\end{figure}

\paragraph{Ablation study for the Black-box generation attack} We study how the number of generated samples affects attack performance. Figure \ref{fig:samples+CC gender} shows the results in Chat-Completion mode and gender property; the estimated gender property ratio converges rapidly: with $1000$ generated samples, the mean absolute error (MAE) drops below $4\%$ for all three model architectures, indicating the attack's efficiency even number limited query budgets. 

Moreover, we study the attack performance with each individual prompt for the BB-generation attack in Chat-completion mode. We observe that there is not a single prompt that achieves the best attack performance across different model architectures; instead, aggregating three prompts either achieves the smallest or the second smallest MAE in three model architectures; hence in the main table, we report the attack performance by aggregating three prompts.

\begin{table}[h!]
\centering
\small
\begin{tabular}{llccc}
\hline
\textbf{Model} & \textbf{Prompt} & \multicolumn{3}{c}{\textbf{Chat-Completion Mode}} \\
\cline{3-5}
 &  & \textbf{30} & \textbf{50} & \textbf{70} \\
\hline
\multirow{3}{*}{LLaMA-1}
& Prompt 1 & 3.80{\scriptsize $\pm$1.76} & 5.10{\scriptsize $\pm$2.65} & 3.35{\scriptsize $\pm$3.62} \\
& Prompt 2 & 3.64{\scriptsize $\pm$1.58} & 4.60{\scriptsize $\pm$5.15} & 5.02{\scriptsize $\pm$4.91} \\
& Prompt 3 & \cellcolor{gray!20} 1.26{\scriptsize $\pm$0.61} & \cellcolor{gray!20} 2.75{\scriptsize $\pm$2.65} & \cellcolor{gray!20} 2.30{\scriptsize $\pm$2.41} \\
& Aggregated & \cellcolor{gray!10}  $1.73{\scriptstyle \pm 0.76}$ & \cellcolor{gray!10} $2.64{\scriptstyle \pm 3.33}$ & \cellcolor{gray!10} $3.28{\scriptstyle \pm 3.64}$ \\
\hline
\multirow{3}{*}{Pythia-v0}
& Prompt 1 & 5.03{\scriptsize $\pm$3.21} & 6.59{\scriptsize $\pm$0.23} & 2.50{\scriptsize $\pm$2.12} \\
& Prompt 2 & \cellcolor{gray!20} 3.10{\scriptsize $\pm$2.30} & \cellcolor{gray!20} 3.98{\scriptsize $\pm$2.11} & \cellcolor{gray!20} 3.01{\scriptsize $\pm$3.16} \\
& Prompt 3 & 4.36{\scriptsize $\pm$4.34} & 6.25{\scriptsize $\pm$0.33} & 4.95{\scriptsize $\pm$2.69} \\
& Aggregated & \cellcolor{gray!10} $3.56{\scriptstyle \pm 2.03}$ & \cellcolor{gray!10} $ 5.61{\scriptstyle \pm 0.78}$ & \cellcolor{gray!10} $2.15{\scriptstyle \pm 2.45}$ \\
\hline
\multirow{3}{*}{LLaMA-3}
& Prompt 1 & \cellcolor{gray!10} 1.93{\scriptsize $\pm$1.11} & \cellcolor{gray!10} 1.49{\scriptsize $\pm$1.41} & \cellcolor{gray!10} 1.96{\scriptsize $\pm$1.57} \\
& Prompt 2 & 0.84{\scriptsize $\pm$0.92} & 2.22{\scriptsize $\pm$2.24} & 2.35{\scriptsize $\pm$2.41} \\
& Prompt 3 & 3.12{\scriptsize $\pm$0.42} & 5.16{\scriptsize $\pm$2.35} & 2.76{\scriptsize $\pm$1.42} \\
& Aggregated & \cellcolor{gray!20} $0.61{\scriptstyle \pm 0.77}$ & \cellcolor{gray!20} $1.33{\scriptstyle \pm 1.31}$ & \cellcolor{gray!20} $1.25{\scriptstyle \pm 1.52}$ \\
\hline
\end{tabular}
\caption{Effect of individual prompts on the BB-generation attack. Reported numbers are the Mean Absolute Errors (MAE; $\downarrow$) between the predicted and target ratios. 
We highlight the attack that achieves the smallest and second smallest total MAE across different target properties: darker grey shades indicate the smallest and the lighter grey shades indicate the second smallest. }
\label{tab:prompt_eval}
\end{table}

\subsection{Defenses} \label{susec: defense}
we implemented a simple defense: since the attack depends on model generations, we adjusted the temperature parameter $T$
 in the final softmax layer. Note that $T > 1$
 makes the model's output more balanced among all tokens and $T<1$
 makes the model's output more concentrated on the high-probability tokens.

 We empirically evaluate this defense on the BB-generation attack using LLaMA-3 fine-tuned in CC mode. The table below reports the average predicted ratios inferred by the attack under different temperature settings, given fine-tuning datasets with varying female ratios.

 \begin{table}[h!]
\centering
\caption{Effect of Different $T$ Values on Target Female Ratios}
\begin{tabular}{@{}lcccc@{}}
\toprule
\textbf{} & \textbf{$T = 1$ (no defense)} & \textbf{$T = 0.5$} & \textbf{$T = 1.5$} & \textbf{$T = 3.5$} \\ 
\midrule
Target Female Ratio = 0.3 & 0.355 & 0.200 & 0.415 & 0.4585 \\
Target Female Ratio = 0.5 & 0.516 & 0.447 & 0.542 & 0.5657 \\
Target Female Ratio = 0.7 & 0.6958 & 0.8686 & 0.6453 & 0.6866 \\
\bottomrule
\end{tabular}
\end{table}

We observe that without defense ($T=1$
), the attack predict the target female ratio mostly correctly. When the temperature is altered ($T \neq1$)
, the predicted ratios have larger error, indicating that adjusting decoding settings can serve as a simple yet effective defense against black-box sampling-based attacks. However, such a defense may break down if the adversary is aware of the default temperature setting or has access to the model weights. A deeper investigation into the limitations of this defense, as well as the development of more robust mitigation strategies, is an important direction for future work.


\newpage
\newpage

\end{document}